\documentclass{article}
  
  \usepackage[final]{nips_2016}
  \usepackage{natbib}
  \setcitestyle{numbers}
  \usepackage[utf8]{inputenc} % allow utf-8 input
  \usepackage[T1]{fontenc}    % use 8-bit T1 fonts
  \usepackage{hyperref}       % hyperlinks
  \usepackage{url}            % simple URL typesetting
  \usepackage{booktabs}       % professional-quality tables
  \usepackage{amsfonts}       % blackboard math symbols
  \usepackage{nicefrac}       % compact symbols for 1/2, etc.
  \usepackage{microtype}      % microtypography
  \usepackage{indentfirst}
  \usepackage{amsmath}
  \usepackage{fancyhdr}
  \usepackage{graphicx}
  \usepackage{printlen}
  \usepackage{color}

  \graphicspath{{./figures/}}
  
  \fancypagestyle{equalc}{\fancyhf{}\fancyfoot[R]{* indicates equal contribution}}
  
  \title{Gradient descent revisited via an adaptive online learning rate}
  
  \author{
	Mathieu Ravaut* \\ 
	University of Toronto \\
	27 King's College Circle\\
	Toronto, ON M5S\\
	mravox@cs.toronto.edu 
	\and
  	Satya Krishna Gorti*\\
	University of Toronto\\
	27 King's College Circle\\
	Toronto, ON M5S\\
	satyag@cs.toronto.edu	
}
  
\begin{document}

  \maketitle
  
  \begin{abstract}
    Any gradient descent optimization requires to choose a learning rate. With deeper and deeper models, tuning that learning rate can easily become tedious and does not necessarily lead to an ideal convergence. We propose a variation of the gradient descent algorithm in the which the learning rate $\eta$ is not fixed. Instead, we learn $\eta$ itself, either by another gradient descent (first-order method), or by Newton's method (second-order). This way, gradient descent for any machine learning algorithm can be optimized. 
  \end{abstract}
  
  \thispagestyle{equalc}
  \section{Introduction}
  
  In the past decades, gradient descent has been widely adopted to optimize the loss function in machine learning algorithms. Lately, the machine learning community has also used the stochastic gradient descent alternative, which processes input data in batches. Gradient descent can be used in any dimension, and presents the advantage of being easy to understand and inexpensive to compute. Under certain assumptions on the loss function (such as convexity), gradient descent is guaranteed to converge to the minimum of the function, \emph{with a carefully chosen learning rate} $\eta$. Moreover, stochastic gradient descent has proven to be very efficient even in situations where the loss function is not convex, as is mostly the case with modern deep neural networks. Other methods such as Newton's method guarantee a much faster convergence, but are typically very expensive. Newton's method for instance requires to compute the inverse of the Hessian matrix of the loss function with regards to all parameters, which is impossible with today's hardware and today's deep networks using millions of parameters. \\
  
  The quality of a gradient descent heavily depends on the choice of the learning rate $\eta$. A too high learning rate will see the loss function jumping around the direction of steepest descent, and eventually diverge. While a very low learning rate prevents non-divergence, convergence will be very slow, and the loss function might get stuck in a local minimum. Choosing an ideal learning rate requires an intuition of the problem. Typically, researchers would start by performing a random search or a line-search over a set of different orders of magnitude of learning rates, but this is long and costly. Besides, these methods assume a fixed learning rate over time, as doing one search per iteration would require exponential computation cost. Usually, setting the learning rate to an initial value and gradually decreasing while training is popularly used.\\
  
  In this paper, we propose to automatically find the learning rate at each epoch. We still need to input an initial value, but at each iteration, our model will find a learning rate that optimizes best the loss function at this point of learning. We explore a first-order method and a second-order one to do so, with a strong emphasis on the latter. Our method could be applied to any machine learning algorithm using gradient descent. We show faster initial convergence on a variety of tasks and models, including:
  \begin{itemize}
  	\item Linear regression on the Boston Housing Prices dataset
  	\item Logistic regression on MNIST
  	\item Image classification with neural networks on CIFAR-10 and CIFAR-100
  \end{itemize} 
  
  \section{Related work}
  
  Techniques for improving optimization of machine learning problems have a long history. The most commonly used technique is Stochastic Gradient Descent (SGD), combined with tricks like momentum \cite{polyak1964some}. While using SGD with complex neural network architectures, the non-convex nature of the problem makes it even harder to optimize. 
  
  Recently used alternatives to SGD include Contrastive Divergence \cite{hinton2006training}, Conjugate Gradients \cite{hinton2006reducing}, Stochastic Diagonal Levenberg-Marquardt \cite{lecun1998gradient}, and Hessian-free optimization \cite{martens2010deep}. All these techniques optimize the parameters of the model with the objective of minimizing a loss function. To the best of our knowledge, none of these techniques have been applied to optimize the learning rate at every iteration as we will discuss.
  
  In practice, techniques of the form $\frac{\eta}{\beta+t}$ that decay the learning rate over time are popular. The Adagrad method \cite{duchi2011adaptive} for instance divides the learning rate at each step by the norm of all previous gradients. Adadelta \cite{zeiler2012adadelta} is an extension of Adagrad that seeks to reduce its aggressive, monotonically decreasing learning rate. Adam \cite{kingma2014adam} is built on top of these methods. It stores an exponentially decaying average of past squared gradients like Adadelta, and also keeps an exponentially decaying average of past gradients, similar to momentum.
  
  LeCun et al. \cite{lecun1998gradient} mention that the optimal learning rate is $\eta_{opt} = \frac{1}{\lambda_{max}}$ where $\lambda_{max}$ is the maximum eigenvalue of the Hessian matrix of the loss function with regards to all parameters. With today's neural network architectures, computing the Hessian is a computationally very expensive operation. Lecun et al. avoid computing the Hessian via approximating its eigenvalues while Martens et al. \cite{martens2010deep} estimate the Hessian via finite differences.
  
  Tom Schaul et al. \cite{schaul2013no} model the expected loss after a SGD update to deduce the best learning rate update \emph{for each dimension}. To do so, they assume a diagonal Hessian matrix. In practice, this is evidently not the case, and they rely on observed samples of the gradient to update the learning rate.
  
  Similar to Martens et al. \cite{martens2010deep} we use finite differences to approximate first order and second order derivatives of the loss function for updating the learning rate at every iteration.

  \section{Adaptive learning rate}
  
  SGD with an adaptive learning rate has the following form:\\
  $$
  \left\{
  \begin{array}{lll}
  w(t+1) & = & w(t) -\eta(t)g(t)\\
  \eta(t+1) & = & h(\eta(t))
  \end{array}
  \right.
  $$
  where $g(t) = \nabla L(w(t))$, $L$ is the empirical loss function, $w(t)$ represents the state of the model's weights at time $t$, and $h$ is a \emph{continuous} function. In the following section we present a first-order and second-order method to update $\eta(t)$.
  
  \subsection{First-order method}
  
  The first-order method consists of doing gradient descent on the learning rate.\\ 
  Let us introduce a function that will be useful in the following:\\
  \begin{align}
  f : R^{n} &\rightarrow R\\
  \eta &\rightarrow L(w(t)-\eta g(t))
  \end{align}
  
  The intuition for the above function is simple: at time $t$ using $\eta(t)$, we suffer a loss of $L(w(t) - \eta(t)g(t))$ in the next iteration, so $f$ represents what the loss would be if we were to perform a gradient descent update with the given learning rate $\eta$.\\
  The first-order method is written as:\\
  $$
  \left\{
  \begin{array}{lll}
  w(t+1) & = & w(t) -\eta(t)g(t)\\
  \eta(t+1) & = & \eta(t) - \alpha f'(\eta(t))
  \end{array}
  \right.
  $$
  This method introduces a new "meta" learning rate $\alpha$. This method is computationally not expensive as we only require to compute one extra gradient of the loss function L as shown in the equation below. \\
  \begin{equation}
  \forall \eta, f'(\eta) = -g(t)^{T}.\nabla L(w(t)-\eta g(t)) \text{ where . is the dot product in dimension $n$}
  \end{equation}
  We can rewrite the above equation as: \\
  \begin{equation}
  f'(\eta(t)) = -g(t)^{T}.g(t+1)
  \end{equation}
  
  The intuition behind this gradient is easy: if we continue in a similar direction then we increase the learning rate, if we backtrack then we decrease it. One
  problem is that the algorithm is not scale invariant anymore, it will behave
  differently for $L^{'}(w) = \lambda L(w)$.
  
  In our experiments, we computed these gradients using finite-difference in places where it was more convenient.
  
  \subsection{Second-order method}
  
  The previous method presents the problem of choosing another meta-learning rate for optimizing the actual learning rate. To avoid this problem, we can use a second-order Newton-Raphson optimization method. \\
  $$
  \left\{
  \begin{array}{lll}
  w(t+1) &= w(t) -\eta(t)g(t)\\
  \eta(t+1) &= \eta(t) - \frac{f'(\eta(t))}{f''(\eta(t))}
  \end{array}
  \right.
  $$
  Now the learning only depends on the loss, and we get rid of the meta-learning rate. \\
  However, the second derivative of $f$ requires building the loss Hessian matrix $H_{L}$:\\
  \begin{equation}
  \forall \eta, f''(\eta) = -g(t)^{T}.H_{L}(w(t)-\eta g(t))
  \end{equation}
  We propose to approximate this Hessian using finite differences similar to \cite{martens2010deep}. By using this approach on $f'$ then $f''$, we get an update formula of the learning rate depending only on several values of the loss function:\\
  \begin{equation}
  f'(\eta+\epsilon) \approx \frac{f(\eta + 2\epsilon)-f(\eta)}{2\epsilon}
  \end{equation}
  \begin{equation}
  \text{and }f'(\eta-\epsilon) \approx \frac{f(\eta)-f(\eta-2\epsilon)}{2\epsilon}
  \end{equation}  
  \begin{equation}
  \text{so }f''(\eta) \approx \frac{f(\eta+2\epsilon)+f(\eta-2\epsilon)-2f(\eta)}{4 \epsilon^{2}}
  \end{equation}
  Given the finite differences on the first derivative:\\    
  \begin{equation}
  f'(\eta) \approx \frac{f(\eta+\epsilon)-f(\eta-\epsilon)}{2 \epsilon}
  \end{equation}
  We get the final simple formula for $\eta$:\\
  \begin{equation}
  \eta(t+1) = \eta(t) - 2\epsilon\frac{(f(\eta+\epsilon)-f(\eta-\epsilon))}{f(\eta+2\epsilon)+f(\eta-2\epsilon)-2f(\eta)}
  \end{equation}    
  Assuming a constant sign for the numerator, this formula means the following: when slightly increasing the learning rate corresponds to a lower loss than slightly reducing it, then the numerator is negative. In consequence, the learning rate is raised at this update, as pushing in the ascending direction for the learning rate seems to help reducing the loss. However, reality is more complex as the denominator's sign changes as well.

  \section{Experiments}
  
  \subsection{Practical considerations}
  
  With the second-order method, the denominator in the learning rate update might underflow. To avoid such a situation, we add to the denominator a small smoothing value to prevent underflow:\\
  \begin{align}
  \text{if } (f(\eta+2\epsilon)+f(\eta-2\epsilon)-2f(\eta)) \approx 0\\
  \text{then }\eta(t+1) = \eta(t) - 2\epsilon\frac{(f(\eta+\epsilon)-f(\eta-\epsilon))}{f(\eta+2\epsilon)+f(\eta-2\epsilon)-2f(\eta)+\delta}
  \end{align}
  A typical value of $\delta$ is $10^{-6}$. \\
  
  Besides, when updating the loss and the learning rate, one should be very careful about the order of the operations. At the \emph{k-th} iteration, we have a loss value $L^{(k)}$ and a learning rate value $\eta^{(k)}$. Then the \emph{(k+1)-th} step of our algorithm is written as follows:\\
  \begin{itemize}
  	\item Get the five loss values $f(\eta^{(k)}+\epsilon)$, $f(\eta^{(k)}-\epsilon)$, $f(\eta^{(k)}+2\epsilon)$, $f(\eta^{(k)}-2\epsilon)$ and $f(\eta^{(k)})$
  	\item $L^{(k+1)} \leftarrow f(\eta^{(k)})$
  	\item $\eta^{(k+1)} \leftarrow \eta^{(k)} -2\epsilon\frac{(f(\eta^{(k)}+\epsilon)-f(\eta^{(k)}-\epsilon))}{f(\eta^{(k)}+2\epsilon)+f(\eta^{(k)}-2\epsilon)-2f(\eta^{(k)})}$
  \end{itemize}

  Note that each update of the learning rate requires five forward passes in the network to get each of the loss values. Theses forward passes are then "canceled" so that the network can come back to its standard state of the beginning of step k. This cancellation is our algorithm's trickiest part to implement.\\
  
  Finally, both first-order and second-order updates of the learning rate don't guarantee that it does not reach negative values. This would be inconvenient as a gradient ascent does not reduce the loss. Therefore, we force the learning  rate to stay positive after each update:\\
  \begin{equation}
  \eta(t+1) = max(\eta(t+1),\delta) \text{ where $\delta$ is the smoothing value}
  \end{equation}
  
  In practice, the learning rate almost never reaches $\delta$ with the second-order method, but this can happen sometimes with the first-order method. 
  
  \subsection{Results} 
  
  We compare our second-order method to a standard SGD method, which uses a fixed learning rate schedule. In the following, this standard method will be called the \emph{basic method}. That means, the basic method uses as optimizer plain stochastic gradient descent with a fixed learning rate. The first order method was quite unstable and less efficient. When relevant, we also include its results. We compare loss and accuracy on both training and test sets. Our goal in these experiments is not to break state of the art on the given problem, but to show an improvement in our method compared to the basic approach, especially in the first few dozens passes through the training dataset. Thus, we do not use any particular trick (not even momentum on the weights update) to improve training accuracy (like data augmentation or parameter tuning), except when we add dropout to prevent overfitting in some cases.
  
  In the plots for all the results, all legends with suffix \textbf{\_basic} refer to simple SGD, and all legends with suffix \textbf{\_adaptive} refer to our adaptive SGD algorithm.
  
  \subsubsection{Linear regression}
  
  We first tried our adaptive learning rate schedule on linear regression applied to the Boston Housing dataset \cite{boston}. This dataset of 506 points with 13 features is suitable for a simple linear regression model. We split this dataset into 400 points for training, and the remaining 106 for test.
  
  The model consists in fitting a vector of parameters $W$ to the dataset. We are minimizing the square loss $L(W)=||X.W-Y||^{2}$.
  
  Given the size of the dataset, we perform gradient descent with a full batch. After line searches, we decided to use $\eta=10^{-6}$ for gradient descent, $\eta_{ini} = 10^{-7}$ and $\alpha = 10^{-9}$ for first-order and $\eta_{ini}=10^{-2}$ for second-order.

  \begin{figure}[!h]
	\includegraphics[width=200pt]{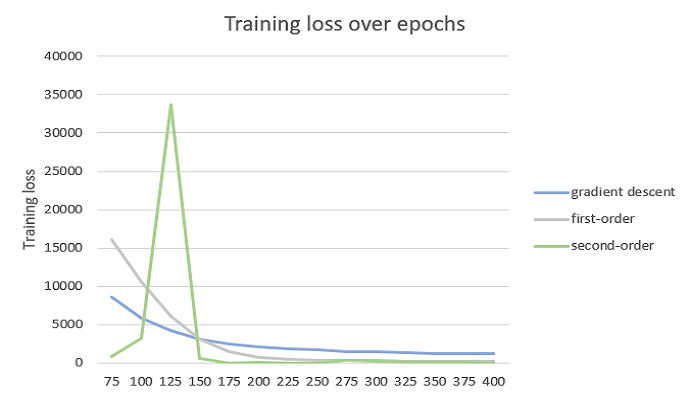}
	\includegraphics[width=200pt]{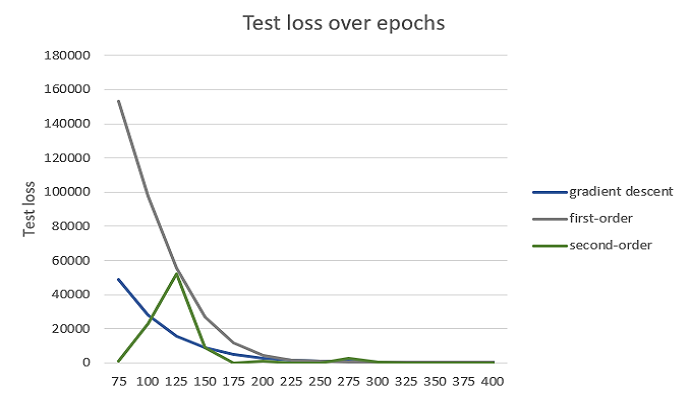}
	\caption{Basic, first-order, and second-order methods training and test losses, Boston Housing}
  \end{figure}

  We can also look at the variations of the learning rate over time for both adaptive methods:\\
  
  \begin{figure}[!h]
	\includegraphics[width=200pt]{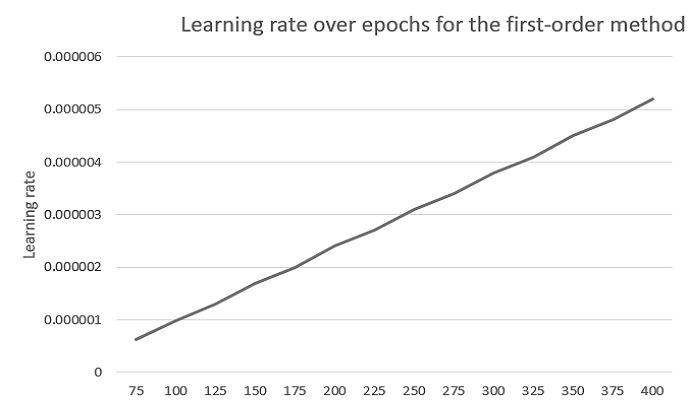}
	\includegraphics[width=200pt]{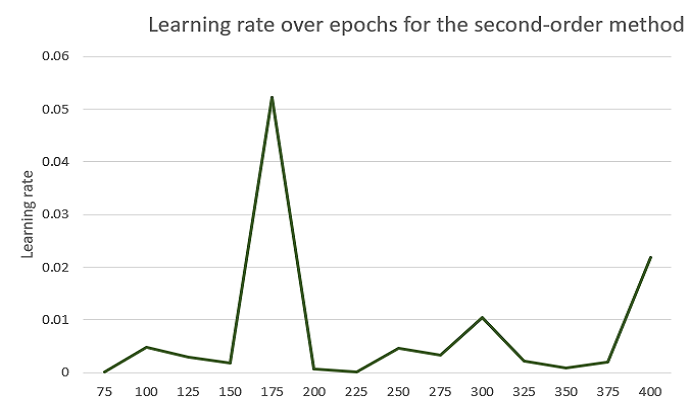}
	\caption{Learning rate for both first-order and second-order methods, Boston Housing}
  \end{figure}  
  
  From these first results, it seems that both the first-order and second-order methods are more efficient than a plain gradient descent. Indeed they reach lower training and test losses, after a comparable time of training, if not faster. However, in this case the second-order method presents some sudden and extreme amplitude variations, that we cannot really explain. There was an extremely high amplitude peak at 175 iterations, that we set down to 0 in the above graphs. Note that when training and test losses spike at 125 iterations, the second-order method reduces its learning rate value soon after.

  \subsubsection{Logistic regression}
  
  We also tried our method with multi-class logistic regression on the MNIST dataset \cite{xiao2017fashion}.
  
  The choice of $\epsilon$ was $10^{-5}$ and we used a meta-learning rate of $\alpha=10^{-6}$ for first-order
  \begin{figure}[!h]
	\includegraphics[width=200pt]{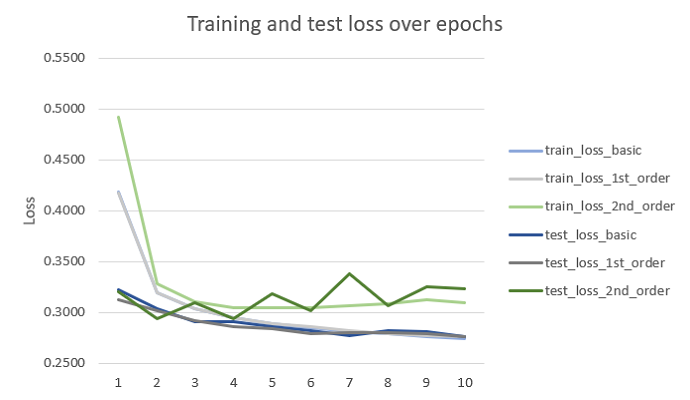}
	\includegraphics[width=200pt]{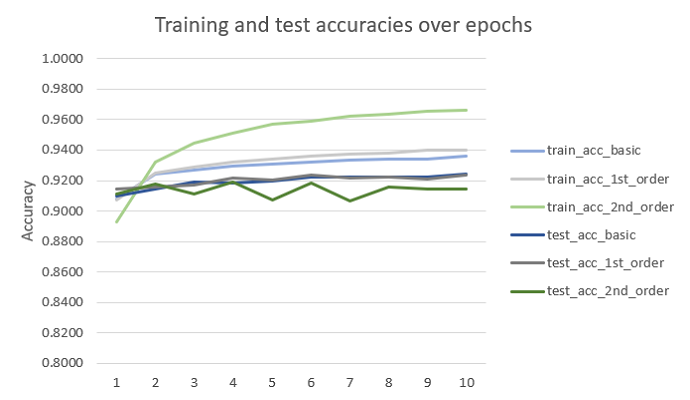}
	\caption{Basic, first-order and second-order losses and accuracies, logistic regression, MNIST}
  \end{figure}  

  \begin{figure}[!h]
	\includegraphics[width=200pt]{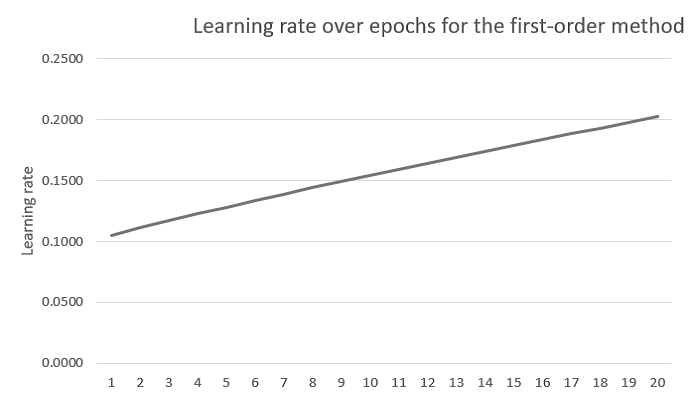}
	\includegraphics[width=200pt]{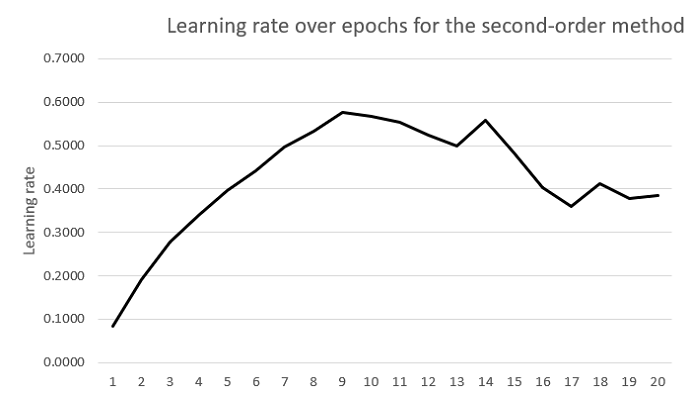}
	\caption{First-order and second-order learning rate variations, logistic regression, MNIST}
  \end{figure} 

  We see that with the second-order method, the learning rate stabilizes around 0.4. 

  \newpage
  \subsubsection{Image classification with neural networks}
  
  \textbf{CIFAR-10}
  
  A common benchmark in machine learning is to classify images on the CIFAR-10 dataset \cite{cifar}. Deep neural networks are most suited for this task \cite{krizhevsky2012imagenet}. In consequence, we used a LeNet with 5 layers (2 convolutional followed by 3 fully-connected layers) and ResNet-18 \cite{he2016deep}. We present results with the second-order method with both networks, then the first-order method with ResNet.
  
  \emph{\textbf{Second-order method}}
  
  We first started by optimizing the LeNet model. We trained it for different learning rate values, and 0.01 appeared to be the best choice. As a plain SGD optimization without momentum tended to overfit, we introduced two dropout layers after the first two fully-connected layers. Thus, we compare the basic method with the second-order adaptive one starting at 0.01, using a batch-size of 32:\\
  
  \begin{figure}[!h]
  	\includegraphics[width=200pt]{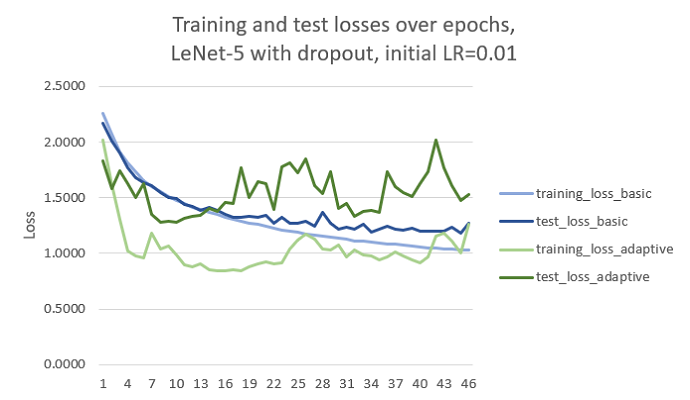}
  	\includegraphics[width=200pt]{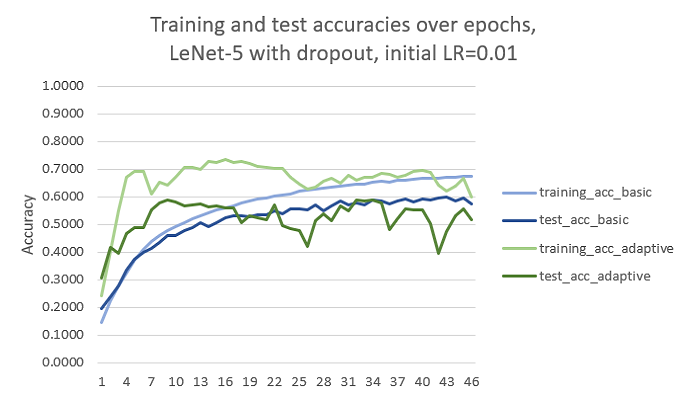}
  	\caption{Second-order and basic methods with LeNet, CIFAR-10, initial learning rate: 0.01}
  \end{figure}

  From here we can notice four things:\\
  \begin{itemize}
  	\item The adaptive method is noisier than simple SGD.
  	\item Training is faster with the adaptive method. Looking at the first epochs shows that the training and test losses get optimized much quicklier than in the basic case. This is of special interest when doing early stopping.
  	\item Given the values that the training loss and accuracy reach, the adaptive method is probably overfitting (but the basic method seems to be on its way to overfit as well).
  	\item Even with this overfitting, the adaptive method reaches the same maximum test accuracy (59\%) than the basic method. This accuracy is reached in less than 10 epochs, versus around 35 epochs for the basic method. 
  \end{itemize}

  We now look at the performance of ResNet-18 (with two added dropout layers). We train it with the reference learning rate of 0.1 given by \cite{he2016deep}, and a batch size of 256:
  
  \begin{figure}[!h]
	\includegraphics[width=200pt]{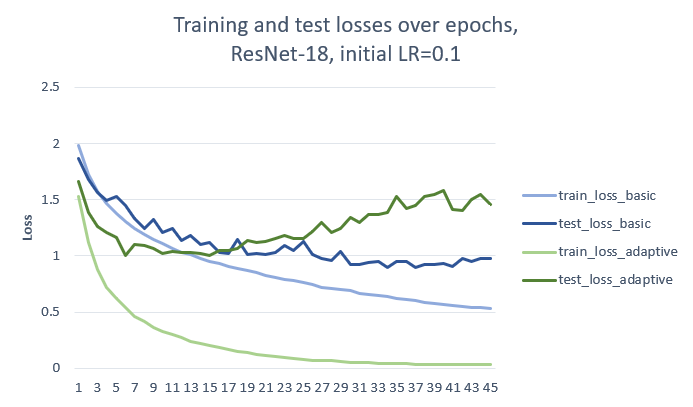}
	\includegraphics[width=200pt]{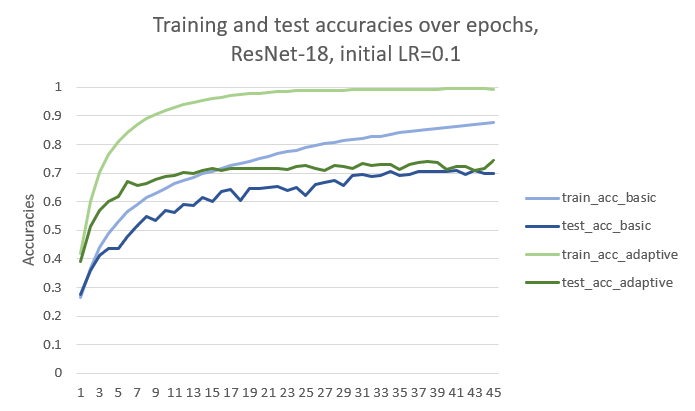}
	\caption{Second-order and basic methods with ResNet, CIFAR-10, initial learning rate: 0.1}
  \end{figure}
  
  Once again, in the adaptive schedule, all training metrics (training and test loss, training and test accuracies) are optimized faster. Typically, in the first epochs, training loss is twice lower in the adaptive version. The second-order method reaches its plateau sooner as well, after around 15 epochs versus 40 in the basic case, and then starts to overfit. Interestingly, this time the best performance achieved by the adaptive method seems to be slightly better (74.4\%) than the best one achieved by the basic method (71\%). In the following, we will try several techniques to improve our method's performance. \\
  This current experiment with ResNet-18 on CIFAR-10 with an initial learning rate of 0.1 will be our reference. We now explore the variations of the learning rate for both networks:
  
  \begin{figure}[!h]
	\includegraphics[width=200pt]{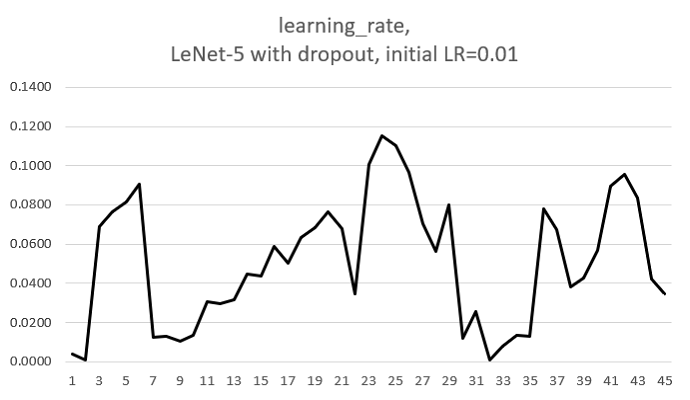}
	\includegraphics[width=200pt]{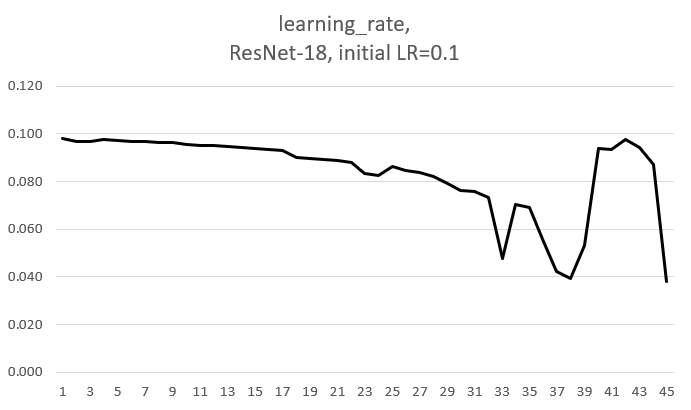}
	\caption{Learning rate variations, LeNet and ResNet, CIFAR-10}
  \end{figure}
  
  For LeNet, it seems that the initial learning rate value was a bit too low. The network increases it to boost convergence, but struggles to stabilize to a fixed zone. With ResNet, it seems that a starting learning rate of 0.1 was a bit too high, as we see it decrease its learning rate. Something interesting starts happening here after 30 epochs, as the learning rate starts oscillating. As the loss value stops changing because convergence is reached, it is probable that the denominator in the learning rate updates gets very small.\\
  
  That leads us to see what happens if we set the initial learning rate at 0.01 for the ResNet model:\\
  
  \begin{figure}[!h]
	\includegraphics[width=130pt]{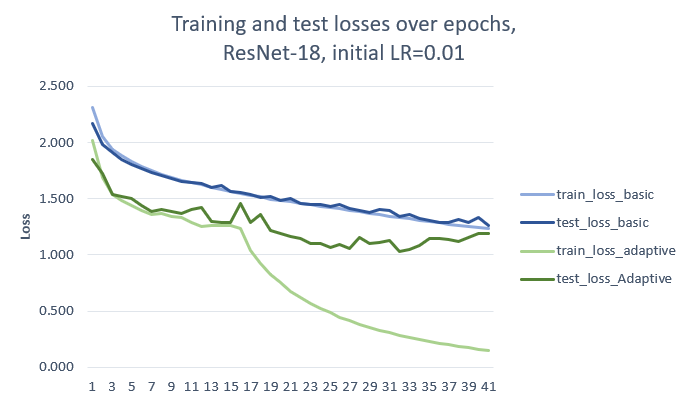}
	\includegraphics[width=130pt]{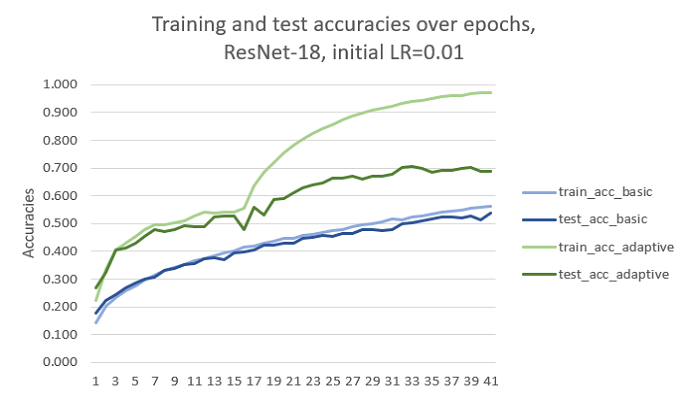}
	\includegraphics[width=130pt]{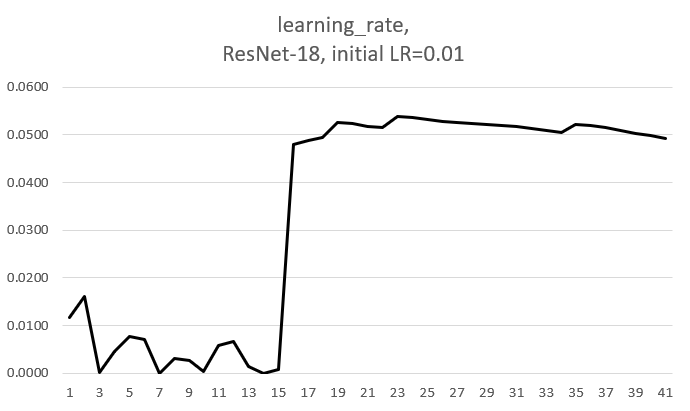}
	\caption{Second-order and basic methods with ResNet, CIFAR-10, initial learning rate: 0.01}
  \end{figure}
  
  The adaptive schedule performs consitently better than the basic one with testing accuracy being 10 to 15 points higher all the way around. This time, the initial learning rate seems to be too small, as the network increases its value until stabilizing it around 0.05, a value around the which it oscillated at the end of the previous training schedule. As training slows down, the network slowly decreases the learning rate value.\\
  
  \emph{\textbf{First-order method}}
  
  \begin{figure}[!h]
	\includegraphics[width=130pt]{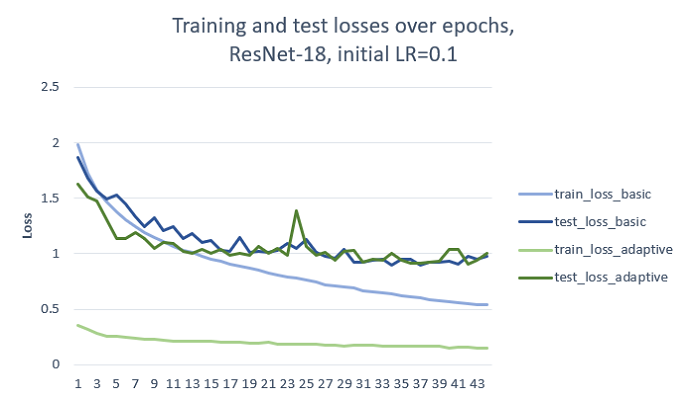}
	\includegraphics[width=130pt]{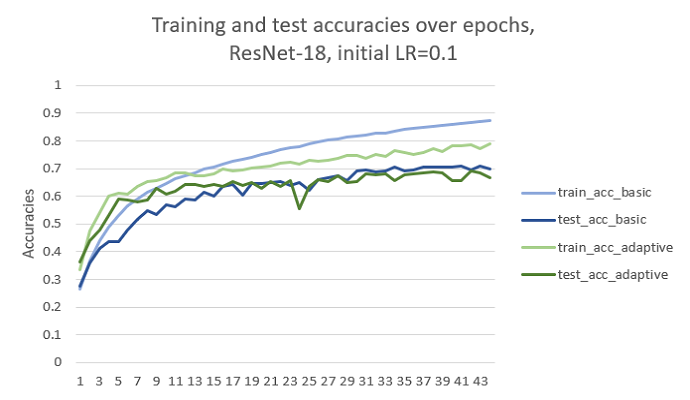}
	\includegraphics[width=130pt]{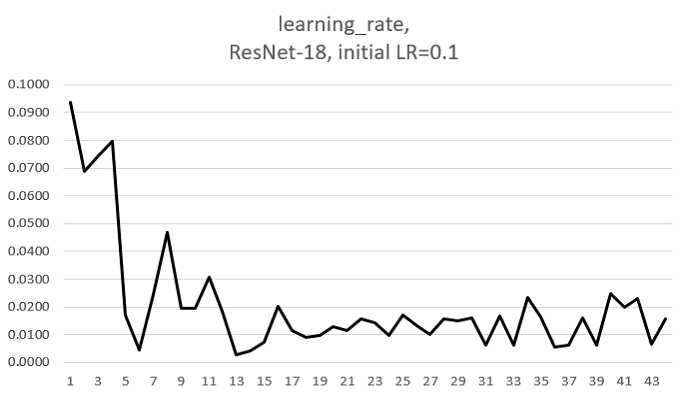}
	\caption{First-order and basic methods with ResNet, CIFAR-10, initial learning rate: 0.1}
  \end{figure}

   The first-order method did not bring much of an improvement in comparison to the reference experiment. The tendency to increase overfitting that we noticed with the second-order method is strengthened here, and the learning rate variations are noisier. However, with lower initial learning rates, the first-order method converges sensibly faster than the basic method, but less than the second-order one. 
  
  \textbf{CIFAR-100}
  
  In this section, we increment difficulty by solving an image classification problem with 10 times more classes (CIFAR-100 \cite{cifar} has 100 classes). After a line-search, the ideal learning rate to train ResNet-18 on CIFAR-100 seems to be 0.02.

  \begin{figure}[!h]
	\includegraphics[width=130pt]{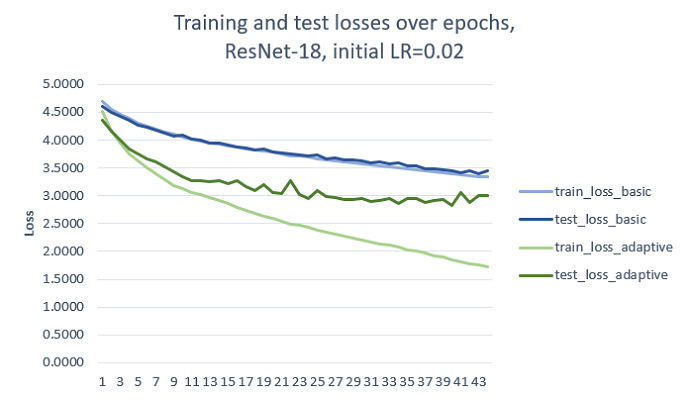}
	\includegraphics[width=130pt]{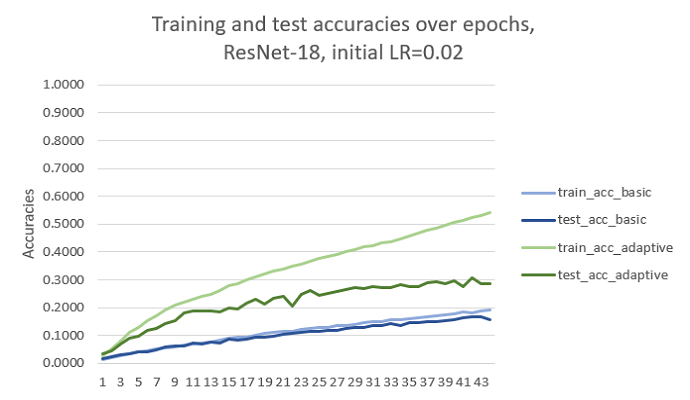}
	\includegraphics[width=130pt]{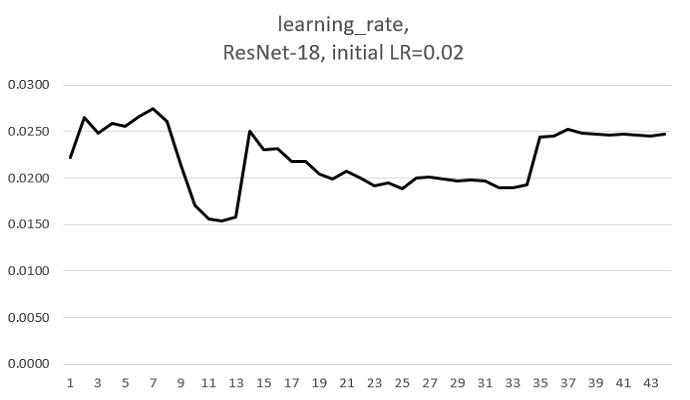}
	\caption{Second-order and basic methods with ResNet, CIFAR-100, initial learning rate: 0.02}
  \end{figure}
  
  Once again, we see a faster loss reduction, a better generalization but also a stronger tendency to overfit. The initial learning rate of 0.02 seemed to be a judicious choice as the network barely changes its value over time. We lacked the computational resource to explore training on this dataset for a longer and more relevant period of time. 
  
  \section{Further exploration}
  
  \subsection{Learning the learning rate}
  
  In the previous experiments, we have seen the learning rate converge to a given value over time (or, at least, stabilize around a certain value). This value seems to depend on both the model and the dataset. Now a question rises: given a dataset and a model, is this the ideal learning rate value to perform gradient descent on this task ?
  
  We started an adaptive learning rate training with the learning rate value that the ResNet model converged to in section 4.2.3 with 0.01 as a starting learning rate. That value is approximatively 0.05
  
  \begin{figure}[!h]
	\includegraphics[width=130pt]{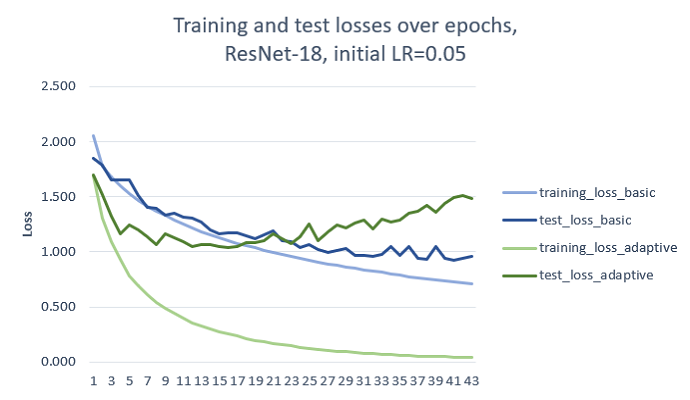}
	\includegraphics[width=130pt]{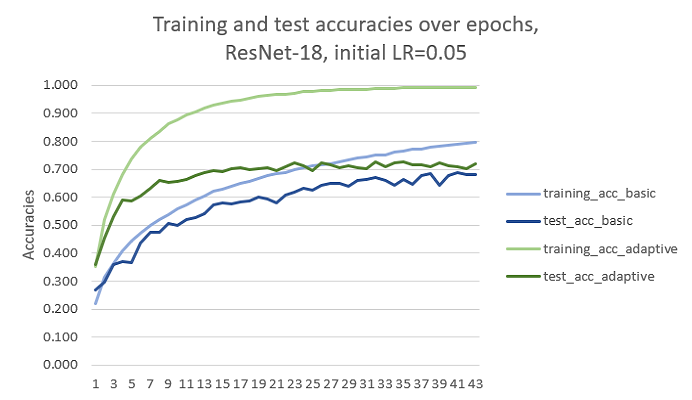}
	\includegraphics[width=130pt]{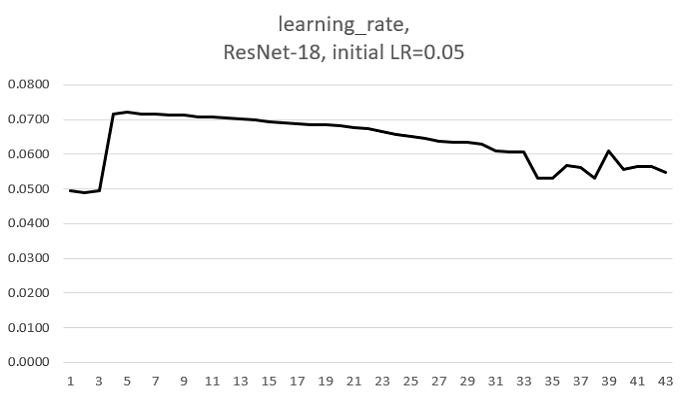}
	\caption{Second-order and basic methods with ResNet, CIFAR-10, initial learning rate: 0.05}
  \end{figure}
  
  The trends noticed in the previous training are confirmed. This time, the learning rate variations are of much weaker amplitude, as this parameter is already in a ideal zone. The network slightly raises its learning rate to kick off training at first, then lowly decreases it over time, as training converges. 
  
  \subsection{Momentum}
  
  In the same way that momentum reduces noise in a gradient descent for weights, we thought that adding momentum on the learning rate updates could help for both our methods. For the second-order method, the update on $\eta$ now is:\\
  $$
  \left\{
  \begin{array}{lll}
  \delta(t+1) &= \beta*\delta(t) - 2\epsilon\frac{(f(\eta+\epsilon)-f(\eta-\epsilon))}{f(\eta+2\epsilon)+f(\eta-2\epsilon)-2f(\eta)}\\
  \eta(t+1) &= \eta(t) + \delta(t+1)
  \end{array}
  \right.
  $$
  
  Here are the results on our reference experiment with $\beta=0.9$:\\
  
  \begin{figure}[!h]
  	\includegraphics[width=130pt]{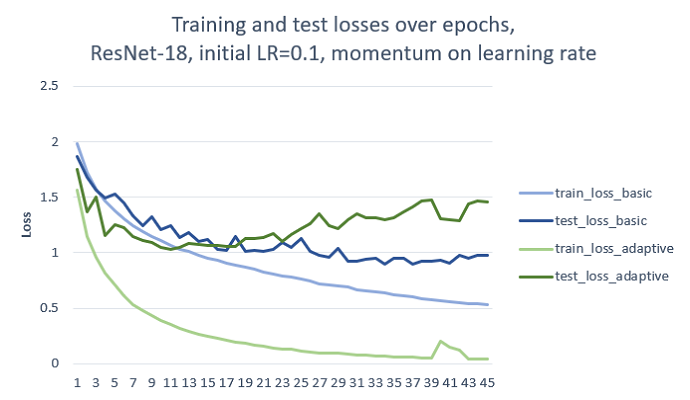}
  	\includegraphics[width=130pt]{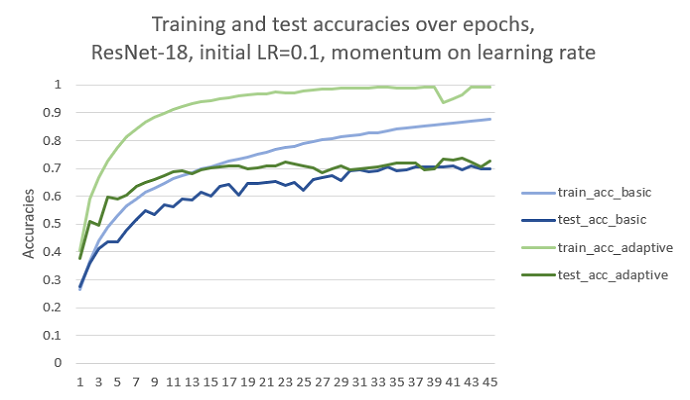}
  	\includegraphics[width=130pt]{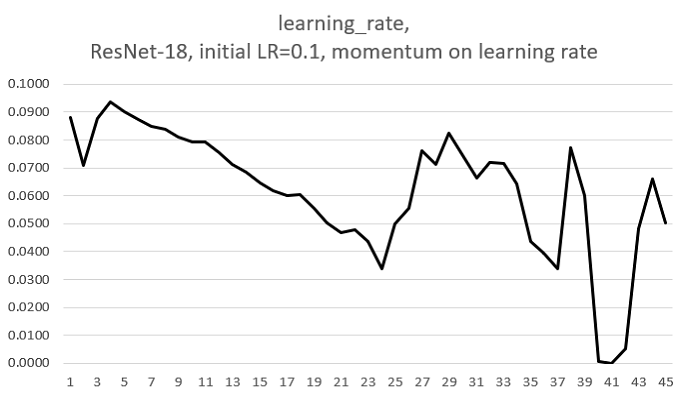}
  	\caption{Second-order with momentum and basic methods with ResNet, initial learning rate: 0.1}
  \end{figure}

  Such a momentum does not seem to bring any particular improvement here. 
  
  \subsection{Getting loss values via a validation set}
  
  One idea to reduce overfitting tendency in the two adaptive methods could be to compute the loss on sample validation batches. This way, the update on the learning rate would guarantee to move in a direction that better reduces the generalization error (and not the training error). The formula for this variant is the same, expect for the use of the $f$ function, that now becomes:\\
  \begin{align}
  f : R^{n} &\rightarrow R\\
  \eta &\rightarrow L_{\text{on a random validation batch}}(w(t)-\eta \nabla L_{\text{current training loss}})
  \end{align}
  
  We partitioned the original training set in a training and validation set, and applied this technique on the reference experiment:\\
  
  \begin{figure}[!h]
	\includegraphics[width=130pt]{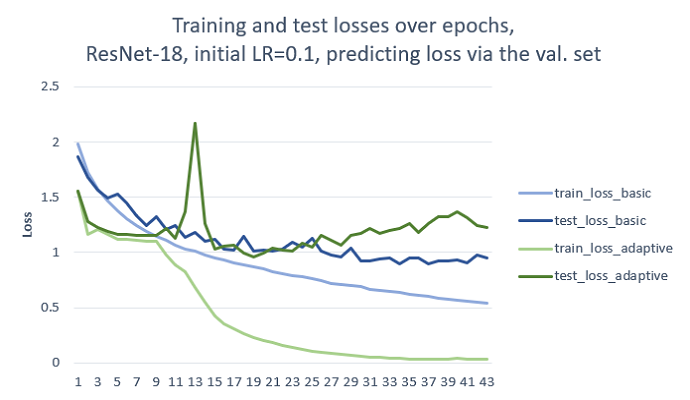}
	\includegraphics[width=130pt]{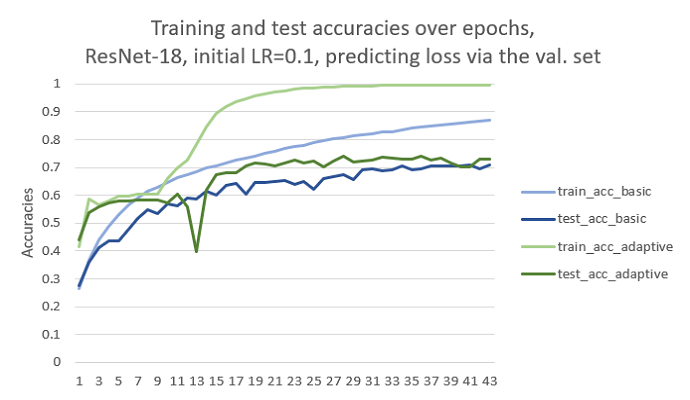}
	\includegraphics[width=130pt]{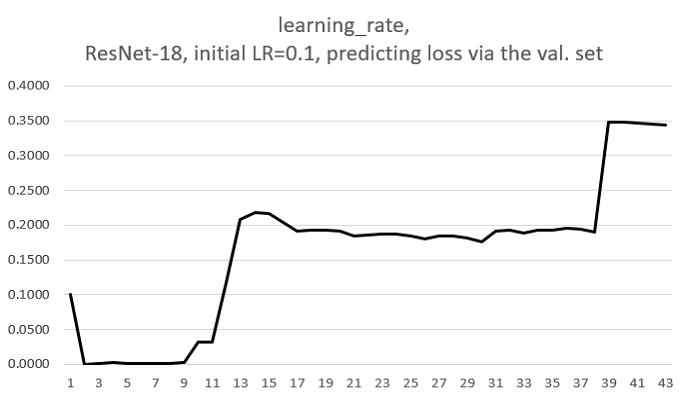}
	\caption{Second-order with val. sampling and basic methods with ResNet, initial learning rate: 0.1}
  \end{figure}

   These results suggest that the training loss and training accuracy get optimized slower, but a deeper investigation would be needed to conclude on a reduction of overfitting.   
  
  \subsection{Comparison with other optimizers}
  
  In the last few years, a popular variation of gradient descent named Adam \cite{kingma2014adam} has progressively gained consensus among researchers for its efficiency. Thus, we compared our method with Adam and changed its default learning rate of $10^{-3}$ to $10^{-1}$, our initial value for the adaptive method.
  
  Here is the comparison with our reference experiment, which results were first shown in section 4.2.3:\\ 
  
  \begin{figure}[!h]
  	\includegraphics[width=200pt]{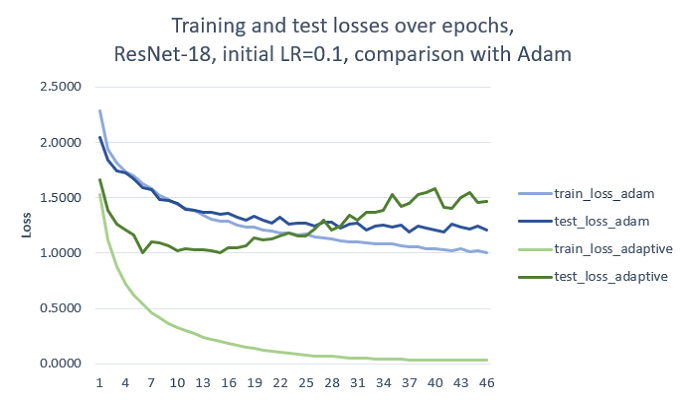}
  	\includegraphics[width=200pt]{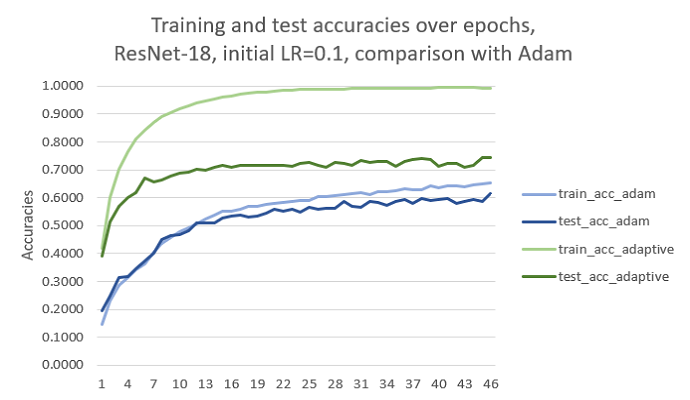}
  	\caption{Second-order and Adam with ResNet, initial learning rate: 0.1}
  \end{figure}

  In the first epochs, our method performs better than an Adam optimization, but the overfitting difference is wider.

  \section{Limitations}
  
  \subsection{Choice of step $\epsilon$}
  
  In finite differences, the choice of the paramater $\epsilon$ is a main issue. Indeed, in any deep learning problem, we can expect the loss function to present noisy oscillations locally. On one hand, with a too small value of $\epsilon$, we might not capture meaningful variations of the $f$ function. On the other hand, a too large $\epsilon$ would make the learning rate oscillate too much, reaching either too high values or even negative values. Both cases can in turn quickly make the loss function diverge. We have seen empirically that when the learning rate gets negative, training will likely not converge. 
  
  All our reported experiments were done using an $\epsilon$ value of $10^{-5}$, which seemed to be a good compromise, and showed stable results for a wide variety of tasks. Tuning $\epsilon$ for a given problem overthrows the interest of not having to tune a learning rate anymore. Ideally, a deeper investigation of finite differences should be able to give values of $\epsilon$ usable for a large variety of tasks. 
  
  \subsection{Choice of the initial learning rate}
  
  In our algorithm schedule, learning rate variations are automatic, but we have to choose the initial value. 
  
  This value seems not to matter so much, as with several ranges of initial values, our algorithm still does better than the basic method. Besides, it always adapts its learning rate to reach the same zone, whatever the initial value (in a reasonable range of learning rate values, e.g. $10^{-4}$ to $1$ for CIFAR-10 for instance).
  
  \subsection{Cost of loss computations}
  
  We have shown faster training in terms of number of epochs. However, at each iteration, we have to perform five forward passes and back-propagations instead of one. Thus, computation time is slowed down compared to the fixed learning rate approach. 
  
  \subsection{Overfitting}
  
  Our version of gradient descent reduces the training loss much faster than in the fixed learning rate approach. The training loss sometimes gets down to very surprisingly low values, while the test loss does not reduce much more than in the basic model. On the contrary, the test loss usually increases and ends at a higher value than the basic method one. Thus, our model seems to be more likely to overfit. We have tried to add dropout to the ResNet model, and that proved successful in reducing the gap between training and testing performance.
  
  However, in terms of accuracy, the second-order method performs systematically as well as the basic method, if not slightly better. Top test accuracy is usually reached early on, which suggests that our adaptive method should be used combined with early stopping.
  
  \section{Conclusion}
  
  In this paper, we have built a new way to learn the learning rate at each step using finite differences on the loss. We have tested it on a variety of convex and non-convex optimization tasks. 
  
  Based on our results, we believe that our method would be able to adapt a good learning rate at every iteration on convex problems. In the case of non-convex problems, we repeatedly observed faster training in the first few epochs. However, our adaptive model seems more inclined to overfit the training data, even though its test accuracy is always comparable to standard SGD performance, if not slightly better. Hence we believe that in neural network architectures, our model can be used initially for pretraining for a few epochs, and then continue with any other standard optimization technique to lead to faster convergence and be computationally more efficient, and perhaps reach a new highest accuracy on the given problem. 
  
  Moreover, the learning rate that our algorithm converges to suggests an ideal learning rate for the given training task. One could use our method to tune the learning rate of a standard neural network (using Adam for instance), giving a more precise value than with line-search or random search.
  
  \bibliographystyle{plain}
  \bibliography{ref}

  \end{document}